\documentclass[final]{cvpr}

\usepackage{times}
\usepackage{epsfig}
\usepackage{graphicx}
\usepackage{amsmath}
\usepackage{amssymb}
\usepackage{caption}
\usepackage{subcaption}
\usepackage{placeins}

\usepackage[pagebackref=true,breaklinks=true,colorlinks,bookmarks=false]{hyperref}

\def\cvprPaperID{40} % *** Enter the CVPR Paper ID here
\def\confYear{CVPR 2021}
\begin{document}

%%%%%%%%% TITLE
\title{Extremely Lightweight Quantization Robust\\ Real-Time Single-Image Super Resolution for Mobile Devices}

\author{Mustafa Ayazoglu\\
Aselsan Research\\
Ankara, Turkey\\
{\tt\small mayazoglu@aselsan.com.tr}
}
\maketitle

%%%%%%%%% ABSTRACT
\begin{abstract}
   Single-Image Super Resolution (SISR) is a classical computer vision problem and it has been studied for over decades. With the recent success of deep learning methods, recent work on SISR focuses solutions with deep learning methodologies and achieves state-of-the-art results. However most of the state-of-the-art SISR methods contain millions of parameters and layers, which limits their practical applications. In this paper, we propose a hardware (Synaptics Dolphin NPU) limitation aware, extremely lightweight quantization robust real-time super resolution network (\textbf{XLSR}). The proposed model's building block is inspired from root modules introduced in \cite{DeepRoots} for Image classification. We successfully applied root modules to SISR problem, further more to make the model uint8 quantization robust we used Clipped ReLU at the last layer of the network and achieved great balance between reconstruction quality and runtime. Furthermore, although the proposed network contains 30x fewer parameters than VDSR \cite{VDSR} its performance surpasses it on Div2K validation set. The network proved itself by winning Mobile AI 2021 Real-Time Single Image Super Resolution Challenge. 
\end{abstract}

%%%%%%%%% BODY TEXT
\section{Introduction}

Super Resolution is a classical computer vision problem and it has been studied over decades. The aim of the problem is obtaining the high resolution image from either single or multiple low resolution images. In both settings the problem is ill-posed. Earlier in the literature the problem is approached with traditional methods later with the advancement and success of deep learning, the problem is approached with deep-learning.

\begin{figure}[h]
\begin{center}
\subfloat[Original: Div2K 0890]{\includegraphics[width=1\linewidth]{"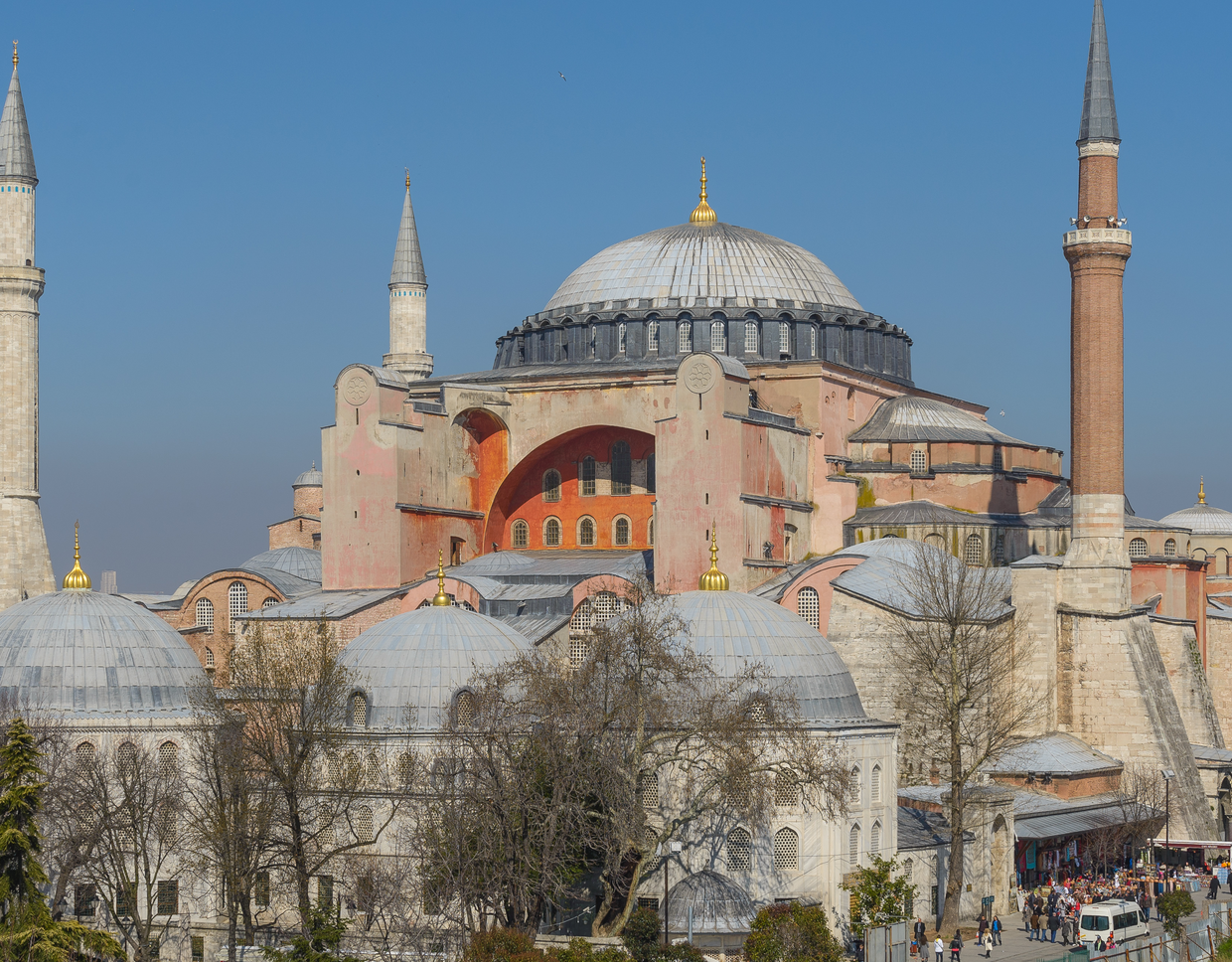"}} \\
\subfloat[\textbf{XLSR (ours)}]{\includegraphics[width=0.33\linewidth]{"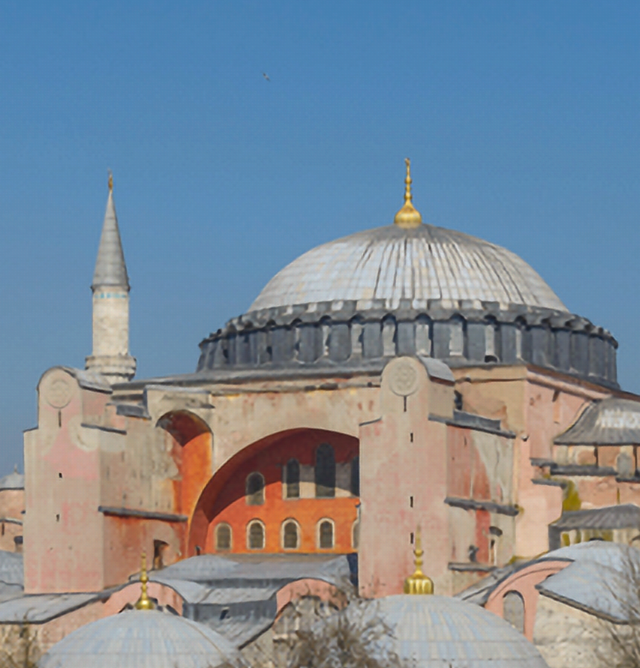"}}
\subfloat[ESPCN]{\includegraphics[width=0.33\linewidth]{"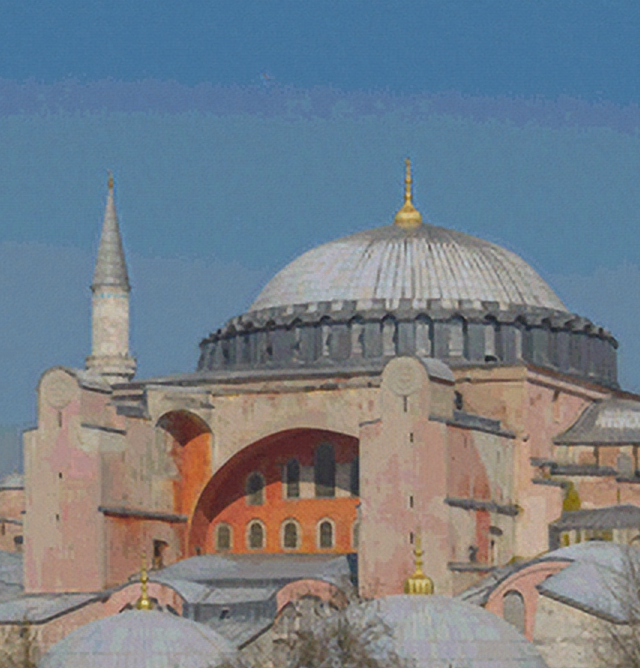"}}
\subfloat[FSRCNN]{\includegraphics[width=0.33\linewidth]{"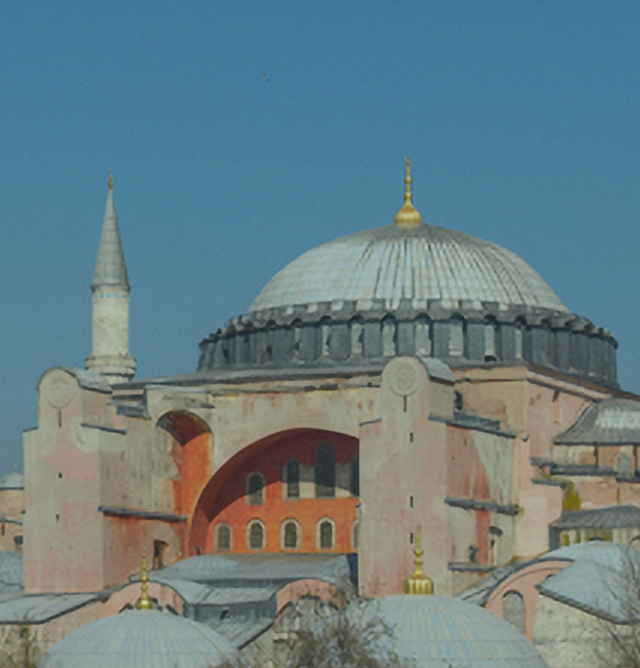"}}
\end{center}
   \caption{Effect of quantization of existing methods and our proposed method. Note the shift in colors and staircase gradient on the sky of quantized ESPCN and FSRCNN outputs}
\label{fig:QuantCompare}
\end{figure}

Deep learning first applied to SISR problem by Dong et al. in \cite{SRCNN} and achieved state-of-the-art results compared to traditional methods with only 3 layers. Researchers realized that upscaling the input image on the later stages of the network reduces computational resources and proposed FSRCNN \cite{FSRCNN} further more instead of ReLU, PReLU is utilized as the activation function, while FSRCNN used deconvolution layer as the upscaling module. Shi et al. \cite{ESPCN} proposed another widely used upscaling layer so called Depth2Space for real-time SISR problem. Later, VDSR \cite{VDSR} emerged and increased the number of layers and number of parameters and showed that with the increased number of parameters the reconstruction quality can be improved. Increasing and complicating the network architecture trend continued later on with EDSR \cite{EDSR} and WDSR \cite{WDSR} later on attention mechanisms applied to SISR in \cite{HAN, IMDN} Indeed these methods achieved superior reconstruction performance on standard datasets. 

While previously mentioned methods focus on reconstruction quality, there are also other methods based on Generative Adversarial Networks (GAN). GAN based methods \cite{SRGAN, ESRGAN} focuses on perceptual quality. 

However, none of these methods and many more in the literature can be directly applied and run in mobile devices either because of extremely large number of parameters and/or severely affected performances due to uint8 quantization, further more some specific hardware limitations may also limit their applicability.

\begin{table}
\begin{center}
\begin{tabular}{|l|c|}
\hline
Method & Number of Parameters \\
\hline\hline
EDSR \cite{EDSR} & 43M \\
WDSR \cite{WDSR} & 75M \\
VDSR* \cite{VDSR} & 668K \\
IMDN \cite{IMDN} & 500K \\
FSRCNN* \cite{FSRCNN} & 25K \\
ESPCN* \cite{ESPCN} & 31K \\
\textbf{XLSR (Ours)} & 22K \\
\hline
\end{tabular}
\end{center}
\caption{Example set of number of parameters from high performing deep learning networks. \\ 
(*) For a fair comparison of parameters model is assumed to accept RGB input and scaling is x3}
\end{table}

\begin{figure*}
\begin{center}
{\includegraphics[clip, trim=3.7cm 4.3cm 2.2cm 5.7cm, scale=0.8]{"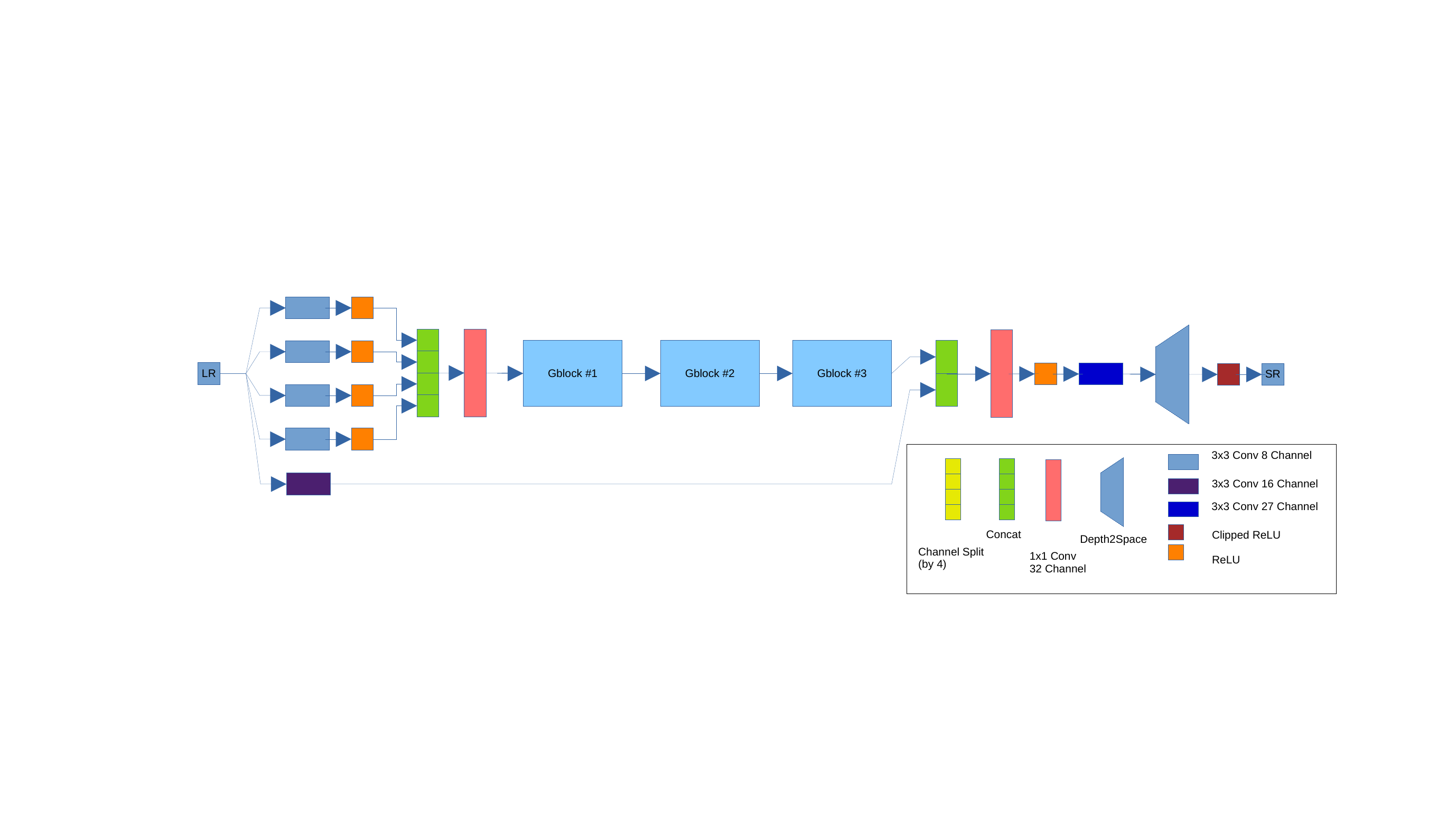"}}
\end{center}
   \caption{Our proposed network}
\label{fig:Network}
\end{figure*}

To solve the real-time image super resolution problem with mobile device deployment requirement, we proposed an extremely lightweight super resolution (\textbf{XLSR}) network by investigating the limiting factors of the existing models to run in mobile devices and modifying successfully applied mobile network building blocks for different problems in the literature (See Figure \ref{fig:Network}). Our method designed with reconstruction quality focus since Mobile AI 2021 Real-time image super resolution challenge scoring formula was based on PSNR as follows;

\begin{gather}
    Score(PSNR,runtime) = \frac{2^{2.{PSNR}}}{C.runtime} \\
    \text{Where \emph{C} is a constant} \nonumber
\end{gather}

Building an extremely lightweight network with low number of parameters was not enough by itself since the model for the challenge needs to be fully uint8 quantized. To make the model quantization robust and keep it still running fast on the deployment platform, instead of "Linear" activation function, which is common at the very last layer of SISR deep learning models, we used "Clipped ReLU" and carefully tuned training methodology. Note that this is required since added non-linearity at the last layer, although helps with quantization, unfortunately makes the model optimization harder due to extra flat optimization surface. It is shown that the model trained with this mentality is very robust to quantization and only $\sim$0.3dB PSNR drop is observed on the quantized model with standard tensorflow post training quantization compared to float16/32 model's PSNR on Div2K validation dataset.

\section{Related Works}

Successful deployment of deep learning models to mobile devices opens broader application areas to these models and increases their usability along with academic contribution. These motivations lead to advancements on both AI specific hardware and on the mobile friendly models. Hardware focusing on deep learning deployment started with the efforts of Qualcomm and Arm later on continued with specialized AI silicon from many different vendors \cite{Ignatov}. On the other hand, many researchers focusing on mobile deployment, come up with many different ideas for mobile friendly models. The examples of these ideas are; for Object detection SqueezeNet \cite{SqueezeNet} which achieves AlexNet \cite{AlexNet} performance on ImageNet with 50x lesser number of parameters. MobileNetV1 \cite{MobileNetV1} uses depthwise separable convolutions and achieves better performance than SqueezeNet. ShuffleNet \cite{ShuffleNet} further increases ImageNet performance with lighter requirements by utilizing group convolutions and channel shuffling operator. Similar to ShuffleNet, DeepRoots proposes the usage of 1x1 convolutions instead of channel shuffling.  For face verification, MobileFaceNets \cite{MobileFaceNets} uses depthwise separable convolutions and bottleneck layers to build an application specific lightweight network. For image super resolution IMDN \cite{IMDN} uses channel splitting to build a lighter network. Different from previous approaches, NASNet \cite{NASNet} searches for the optimal architecture and surpasses many state-of-the-art methods. 

Apart from focusing on network itself, the literature also focuses on quantization \cite{DoReFa, Hubara, PACT, PAMS} since in many cases quantization is not an option but a hardware declared must. On the other hand, it has been shown by Hinton et al. \cite{KnowDistill} that knowledge distillation can help a simple model to achieve/surpass a complex model's performance. Furthermore, Mishra et al. \cite{Apprentice} combined knowledge distilling with quantization. Another methodology using a pretrained complex model while building a lighter weight model is; channel sparsification and pruning, with this methodology a complex model can be slimmed by removing unnecessary channels from the filters \cite{MegviiPrune}

Although the most of the aforementioned methods are designed for different areas other than SISR, they can be still very useful while building a real-time performant SISR model. Many of these ideas are succesfully applied to the SISR problem \cite{Zhang19,Zhang20}  The methodologies to run/design a performant deep learning method for mobile devices can be summarized as follows;

\begin{itemize}
\setlength\itemsep{0.5em}
\item Hand-Designed Architectures
\item Efficient Building Block Design
\item Network Pruning / Sparsification
\item Network Quantization
\item Network Architecture Search (NAS)
\item Knowledge Distillation
\end{itemize}

In this paper, we followed the first and second methodologies while creating our submission into Mobile AI 2021 Real-Time Single Image Super-Resolution Challenge \cite{MAI2021Report}. The reason and motivation for this approach was; there were many different hardware limitations of the deployment platform (Synaptics Dolphin NPU) that needs to be taken into account in the challenge which are harder to incorporate into a common method. Furthermore, the challenge required full uint8 quantization of the model (not allowing partial quantization such as leaving the first and the last layers floating, which are known to be severely affecting the accuracy if quantized \cite{PACT}). The full uint8 quantization requirement adds extra complexity to the problem since, SISR problem is severely affected by the quantization operation if the model at the hand is not designed/trained/quantized properly.

The deployment hardware limiting factors can be summarized as

\begin{itemize}
\setlength\itemsep{0.5em}
\item Elementwise operations such as (Addition, Subtraction) is not optimized
\item Reshaping and transpose operations are not optimized and extremely slow
\item Per-channel quantization is not supported
\item Multiple and especially long skip connections require a lot of data swaps with slow CPU DRAM
\end{itemize}

\section{Proposed Method}

In this section, we describe the details of the proposed network and the motivation behind the design ideas while connecting these with literature and hardware limitations. As mentioned before, we designed our proposed architecture by hand and adopted an efficient building block for SISR problem inspired from \cite{ShuffleNet, DeepRoots, ResNeXt}.

\subsection{Building Block Selection}

Group convolutions are first used in AlexNet although the GPU hardware limitation forced such methodology. It has been shown that when used wisely, group convolutions can increase accuracy while decreasing computational costs. Because of these properties, they are often used in mobile focused networks. They are used in ResNeXt along with skip connections, in ShuffleNet with cascaded channel shuffling and in DeepRoots with cascaded 1x1 convolutions. 

From the point of the view of the SISR problem challenge and hardware limitations, using channel shuffling is infeasible since reshape and transpose operations are not optimized in the deployment hardware which eliminates ShuffleNet Block. Skip connections and residual in residual type of structures help with model converge and allow deeper architectures and are utilized in many state-of-the-art networks. However, multiple parallel skip connections are slow and elementwise addition is not optimized and hence ResNeXt block is not very well optimized for the deployment hardware. 

\begin{figure}[t]
\begin{center}
\includegraphics[clip, trim=8.7cm 5.5cm 12cm 3.8cm, scale=1.2]{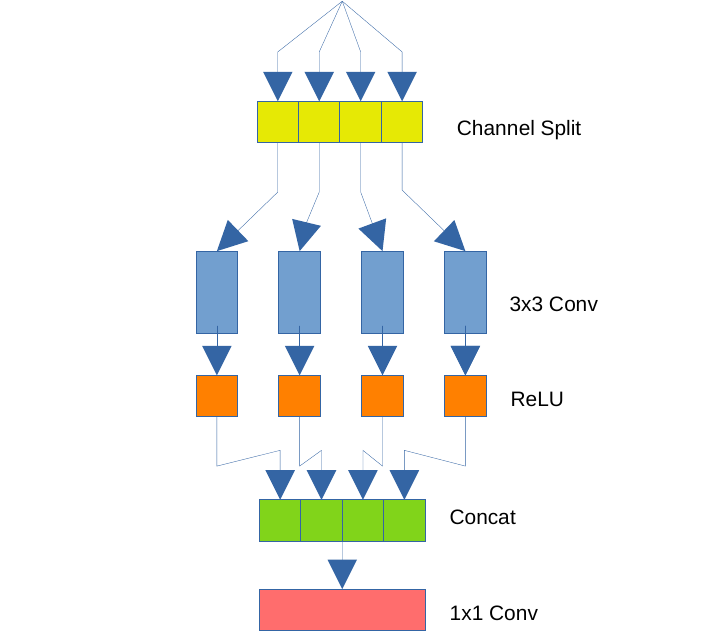}
\end{center}
   \caption{Building block used in the proposed network}
\label{fig:GBlock}
\end{figure}

On the other hand, when channel shuffling operator is relaxed to 1x1 convolution (to still allow interchannel communication) or skip connection in ResNext block is removed, we arrive at the building block used in our network (See Figure \ref{fig:GBlock}). Note that instead of group convolutions, depthwise convolutions can also be a candidate for building block. However, as noted by Sheng et al. in \cite{QualcommQuant} depthwise convolutions can cause large quantization error when used without special precaution, this is what we have also observed empirically in our experiments.

An important aspect we have also taken into account while designing our proposed model was elementwise operations are not optimized in the deployment hardware. Thus, we avoided all addition and scaling operations and used concatenation operation when necessary and input data scaling and normalization were not used.

\subsection{Quantization Friendly Architecture}

\begin{figure*}
\begin{center}

\begin{tabular}{ccc}
\begin{tabular}{@{}c@{}} 
\subfloat[Original: Div2K 0833]{\includegraphics[width=0.45\linewidth]{"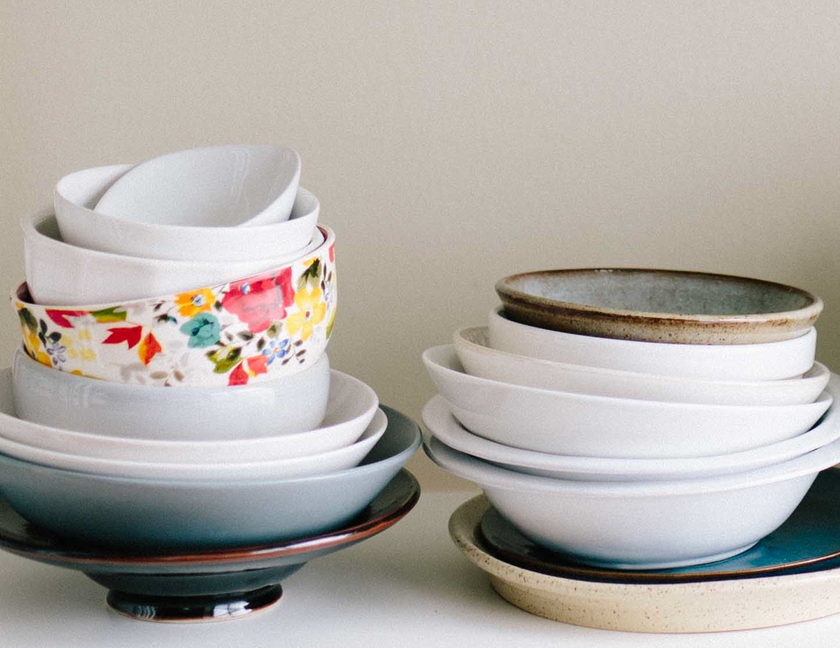"}} \\ 
\end{tabular} & 
\begin{tabular}{@{}c@{}} 
\subfloat[HR]{\includegraphics[width=0.22\linewidth]{"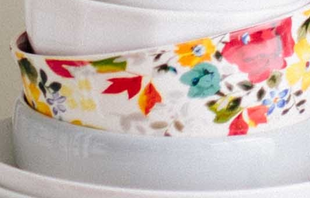"}} \\ 
\subfloat[ESPCN]{\includegraphics[width=0.22\linewidth]{"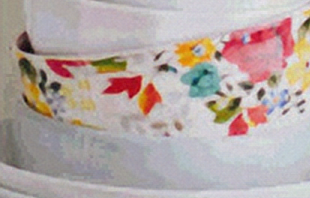"}} 
\end{tabular} & 
\begin{tabular}{@{}c@{}} 
\subfloat[\textbf{XLSR (ours)}]{\includegraphics[width=0.22\linewidth]{"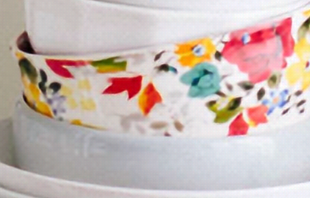"}} \\ 
\subfloat[FSRCNN]{\includegraphics[width=0.22\linewidth]{"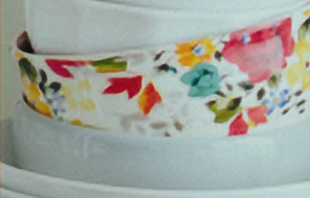"}} 
\end{tabular}
\end{tabular}

\vspace{5mm}

\begin{tabular}{ccc}
\begin{tabular}{@{}c@{}} 
\subfloat[Original: Div2K 0823]{\includegraphics[width=0.45\linewidth]{"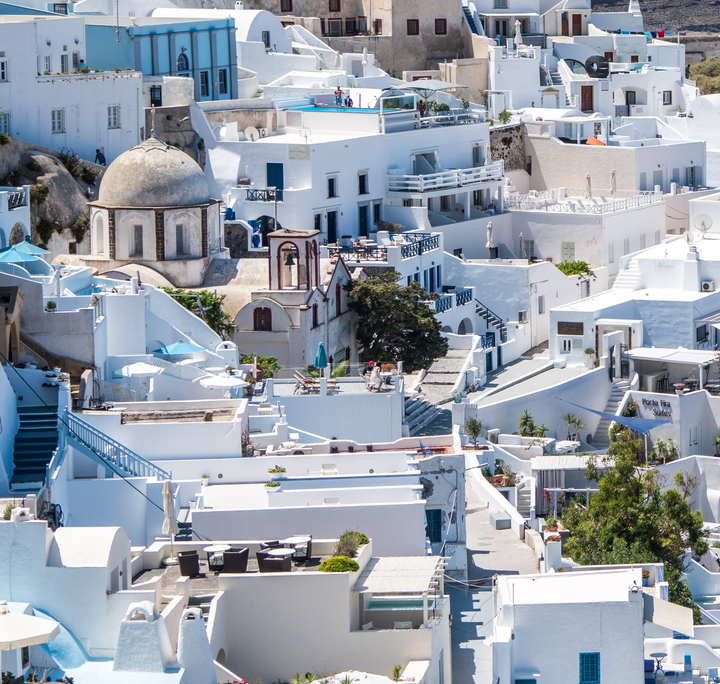"}} \\ 
\end{tabular} & 
\begin{tabular}{@{}c@{}} 
\subfloat[HR]{\includegraphics[width=0.22\linewidth]{"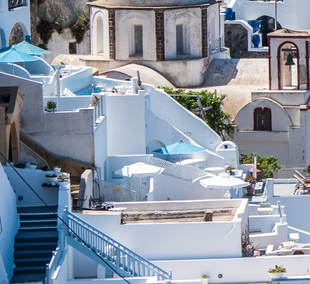"}} \\ 
\subfloat[ESPCN]{\includegraphics[width=0.22\linewidth]{"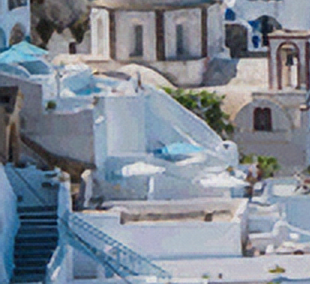"}} 
\end{tabular} & 
\begin{tabular}{@{}c@{}} 
\subfloat[\textbf{XLSR (ours)}]{\includegraphics[width=0.22\linewidth]{"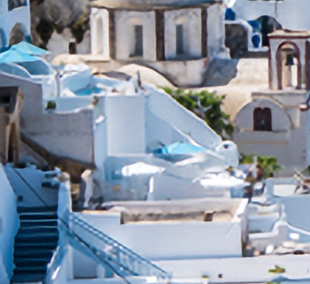"}} \\ 
\subfloat[FSRCNN]{\includegraphics[width=0.22\linewidth]{"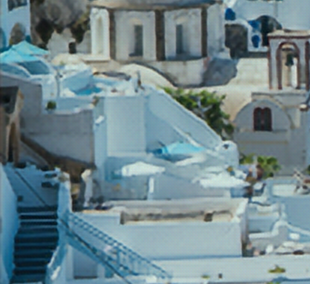"}} 
\end{tabular}
\end{tabular}

\end{center}
   \caption{Example Images from Div2K Dataset. Note false colors and staircase gradient effect on quantized ESPCN and FSRCNN output}
   
\label{fig:Div2k833822}
\end{figure*}

While deploying deep learning models to mobile devices, it is very common to use quantization especially 8-bit quantization is very important since it is naturally supported by many mobile hardware as with the deployment hardware of Mobile AI 2021 Real-time single image super resolution challenge. However, applying uint8 quantization scheme to any well performing (in float32 or float16) super resolution model and hoping to arrive at a well performing integer quantized model is very naive and usually not working well in practice (See Figure \ref{fig:Div2k833822}). To get over this problem and still have a well performing model, the model should be modified and made quantization friendly. On the other hand, this modification should be subtle and still be quickly executable, otherwise it would contradict with real-time inference requirement. 

To make a model quantization friendly, we should first focus on the reasons behind why quantization adversely affects the accuracy. Linear output activation function is very common among super resolution models and it helps with the model optimization. Although, not strictly enforced when the model converges we are pretty sure that the output will be bounded in between 0-255 (or 0-1.0). However, if such a model is quantized, the quantized output tend to be dull and accuracy can drop about 5-7dB compared to float16/32 model. The reason we believe is the following; during the earlier steps of training, the output is not guaranteed to be in between 0-1.0 and intermediate activations can also be unbounded or might contain outliers. Later on, if we continue with training the training data enforces boundedness in the model output indirectly. However, the intermediate activations of the model are not acted upon and the model may converge to a point near where its intermediate activations are unbounded, since model usually visits these points in the early stages of the training. These allowed intermediate unbounded activations create outliers (very large a few numbers). The outliers in intermediate layer activations, when uint8 quantized leads to some important information carrying, comparably low amplitude values to be zeroed out. Hence effective signal energy reaching to the last layer drops, which results in dull colors and drastic PSNR drops.

 With this motivation we believe if the intermediate layer activations are forced not to visit these outlier creating points from the beginning of the training, the model would be more quantization friendly. This is exactly what happens when the model is trained with Clipped ReLU at the last layer. A model trained with this mentality results in only $\sim$0.2-0.5dB loss with respect to its floating counterpart. This is better seen in Table \ref{tab:ClippedReluExp} where we trained our proposed model without Clipped ReLU and quantized the resulting model and noted Div2K \cite{Div2K} validation accuracy. 
 %Sample output images from the Linear Activation and Clipped ReLU activation networks can be seen in Figure \ref{fig:ClippedReluExp}, note the dull and washed out colors with Linear activation.
 
\begin{table}[htp]
\begin{center}
\begin{tabular}{|l|c|c|c|}
\hline
Activation & PSNR (fp32) & PSNR (uint8) & Drop  \\
\hline\hline
Linear & 30.11 & 24.72 & 5.39\\
\textbf{Clipped ReLU} & \textbf{30.10} & \textbf{29.82} & \textbf{0.28}\\
\hline
\end{tabular}
\end{center}
\caption{Effect of Clipped ReLU and Div2K Validation Set PSNR Results with floating and uint8 quantized models}
\label{tab:ClippedReluExp}
\end{table}

%\begin{figure*}
%\begin{center}
%\subfloat[Clipped ReLU]{\includegraphics[width=1\linewidth]{"./images/image_35_clipped.png"}} \\
%\subfloat[Original]{\includegraphics[width=1\linewidth]{"./images/image_35_Original.png"}} \\
%\subfloat[Linear]{\includegraphics[width=1\linewidth]{"./images/image_35_linear.png"}}
%\end{center}
%   \caption{Clipped ReLU vs Linear Activation Experiment Sample Image, Note whites and blacks are on the grayish side when there is no Clipped ReLU at the last layer}
%\label{fig:ClippedReluExp}
%\end{figure*}

Note that Clipped ReLU is
 
\begin{gather}
    Clipped\_ReLU(x) = max(0,min(x,1))
\end{gather}
 
so it is very cheap to calculate. Thus the computational burden added to the model is minimal which is desired.

Furthermore, although placing a single Clipped ReLU was enough to regularize the intermediate activations of our proposed network, when a model gets deeper, regularization effect might vanish for deeper layers, to overcome this issue we suggest to change a few of the ReLU's with Clipped ReLU's. We believe that clipping only a few ReLU's is enough to regularize the intermediate activations though we have not experimented on this idea for this paper. Also note that for intermediate Clipped ReLU's clipping value does not need to be 1 and it should either be experimentally found or it should be included in the optimization as done in \cite{PAMS}.
 
 One drawback of using Clipped ReLU at the output layer is unfortunately the model is harder to optimize since it creates a lot of flat regions with minimal or no derivative direction. Because of this reason, to converge to a better model, a few training tricks needs to be employed which are explained in training details.

\section{Experiments}
\subsection{Datasets}
We used the supplied Div2K Dataset \cite{Div2K} for the challenge and no extra data is used. The dataset consist of 800 high quality training images and 100 validation images along with 100 test images. During the challenge the test images were not released. Due to this, while reporting our results we only used validation set results (which are not used either for training or validation). For a fair comparison, on standard benchmark datasets (Set5, Set14, BSD100, Manga109, Urban100) with the existing work, we used our floating point model and used Y channel of our output.

\subsection{Training Details}
 For training, we splitted 800 training images into two sets, we used the first 792 of training images for training while keeping last 8 images for validation. We cropped random 32x32 LR images and for data augmentation we used geometric transformations (8 transformations - original, rotations, flips) with equal probability. Furthermore, to give more robustness to illumination changes we randomly scaled the intensity of images with 1, 0.7 and 0.5 randomly. As the loss function, we used Charbonnier loss \cite{Charbonnier} with \( \epsilon = 0.1\) as defined in (\ref{mat:charbon}) since it is the smoother version of L1 and we empirically found out that it works better with Clipped ReLU
 
\begin{gather}
    Charbonnier(x) = \sqrt{x^2+\epsilon^2}
\label{mat:charbon}
\end{gather}

As mentioned above, the critical part of the model development was to include Clipped ReLU activation function. This is however on the other hand results in a harder problem. So to still be able to converge to a performant model following tricks were utilized;

\begin{itemize}
\setlength\itemsep{0.5em}
\item We used a triangular cyclic learning rate scheduling strategy \cite{Leslie}, which starts with a very low value (5e-5) for the first epoch and quickly increases to the top value(25e-4) in 50 epochs and slowly decreases to a low value till 5000 epochs (1e-4). 
\item Trained the model with mini batch size 16 for 5000 epochs with each epoch containing 100 mini batches. At the end of each epoch, we calculated the validation set PSNR (last 8 images of training set) and saved the best model for quantization
\item Used Adam optimizer with default beta1=0.9 and beta2=0.999 and epsilon=1e-8 values
\item Initialized Conv2D layers with He-Normal with 0.1 Variance Scaling \cite{EDSR} The motivation behind this was to initialize kernels to closer to zero and avoid large numbers which might in turn create outliers in activation
\end{itemize}

For quantization, we used  standard Tensorflow Post-Quantization strategy and trained the model on NVIDIA RTX 2080 Super. It takes about 2 hours to train our proposed model in that hardware.

Our quantized model achieves the following runtimes in Samsung Galaxy A21s using AI-Benchmark Tool \footnote{https://ai-benchmark.com/download} \cite{Ignatov};

\begin{itemize}
\setlength\itemsep{-0.2em}
\item CPU Runtime ~1130ms
\item NNAPI Runtime ~1150ms
\item GPU Delegate Runtime ~620ms
\end{itemize}

\textbf{XLSR} also achieves an exceptional target platform runtime of 45ms. Table \ref{tab:MAIresults} shows the results from Mobile AI 2021 Real-Time Super Resolution Challenge \cite{MAI2021Report}. 

\begin{table}
\begin{center}
\begin{tabular}{|l|c|c|c|c|}
\hline
Method & PSNR & SSIM & Runtime & Score \\
\hline\hline
\textbf{XLSR} (ours) & 29.58 & 0.86 & 44.85ms & 51.02 \\
2nd Method & 29.41 & 0.8537 & 38.32ms & 47.18 \\
3rd Method & 29.52 & 0.8607 & 62.25ms & 33.82 \\
4th Method & 28.82 & 0.8428 & 76.61ms & 10.41 \\
5th Method & 28.92 & 0.8486 & 718.32ms & 1.28 \\
\hline
\end{tabular}
\end{center}
\caption{Mobile 2021 Real-Time Single Image Super Resolution Results and Scores on Div2K Test Dataset (not released)}
\label{tab:MAIresults}
\end{table}

\begin{table*}
\begin{center}
\begin{tabular}{|l|c|c|c|c|c|c|c|c|}
\hline
Dataset & Scale & Bicubic & FSRCNN & ESPCN & VDSR & IMDN & EDSR & \textbf{XLSR (ours)} \\
\hline\hline
Set5 & x3 & 30.41 & 33.16 & 33.13 & 33.66 & 34.36 & 34.65 & 33.42\\
Set14 & x3 & 27.55 & 29.43 & 29.42 & 29.77 & 30.32 & 30.52 & 29.73\\
B100 & x3 & 27.22 & 28.60 & 28.50 & 28.82 & 29.09 & 29.25 & 28.55 \\
Urban100 & x3 & 24.47 & 26.48 & 26.41 & 27.14 & 28.17 & 28.80 & 26.71 \\
Manga109 & x3 & 26.99 & 30.98 & 30.79 & 32.01 & 33.61 & - & 31.63 \\
\hline\hline
Div2K(Val) & x3 & 28.22 & - & - & 30.09 & - & 31.26 & \textbf{30.10}\\
\hline
\end{tabular}
\end{center}
\caption{Comparison of our proposed method with public benchmark scores of other methods. Note that for fair and consistent comparison with the literature, we used our floating point model and and converted RGB output of our method to Luminance (Y) channel only while calculating PSNR for all dataset except Div2K}
\label{tab:floatPSNR}
\end{table*}

PSNR results of our proposed method, compared with some of the work in the literature on the standard benchmark dataset and Div2K validation set can be seen in Table \ref{tab:floatPSNR}. Note that for a fair and consistent comparison with the literature, the result presented here are from our float32 model. Comparison of our quantized model with quantized FSRCNN and ESPCN can be found in Table \ref{tab:uint8PSNR}.

\begin{table}
\begin{center}
\begin{tabular}{|l|c|c|c|}
\hline
Model  & FSRCNN* (dB) & ESPCN* (dB) & XLSR (dB) \\
\hline\hline
float32 & 29.67 & 29.54 & \textbf{30.10}\\
uint8 & 21.95 & 23.93 & \textbf{29.82}\\
drop & 7.72 & 5.61 & \textbf{0.28}\\
\hline
\end{tabular}
\end{center}
\caption{Effect of quantization on the existing methods and our method on Div2K validation (x3 scale) set.
(*) Our implementation and training of the models to convert them to RGB input and RGB output}
\label{tab:uint8PSNR}
\end{table}

%\begin{figure*}
%\begin{center}
%\subfloat[Original]{\includegraphics[width=0.5\linewidth]{"./images/image_32_Original.png"}}
%\subfloat[XLSR]{\includegraphics[width=0.5\linewidth]{"./images/image_32_XLSR.png"}} \\
%\subfloat[ESPCN]{\includegraphics[width=0.5\linewidth]{"./images/image_32_ESPCN.png"}}
%\subfloat[FSRCNN]{\includegraphics[width=0.5\linewidth]{"./images/image_32_FSRCNN.png"}} \\
%\end{center}
%   \caption{Div2K Image No: 0833. Note false colors and staircase gradient effect on quantized ESPCN and FSRCNN output}
%\label{fig:Div2k833}
%\end{figure*}

%\begin{figure*}
%\begin{center}
%\subfloat[Original]{\includegraphics[width=0.5\linewidth]{"./images/image_22_Original.png"}}
%\subfloat[XLSR]{\includegraphics[width=0.5\linewidth]{"./images/image_22_XLSR.png"}} \\
%\subfloat[ESPCN]{\includegraphics[width=0.5\linewidth]{"./images/image_22_ESPCN.png"}}
%\subfloat[FSRCNN]{\includegraphics[width=0.5\linewidth]{"./images/image_22_FSRCNN.png"}} \\
%\end{center}
%   \caption{Div2K Image No: 0867. Note false colors and staircase gradient effect on quantized ESPCN and FSRCNN output}
%\label{fig:Div2k867}
%\end{figure*}

%clip, trim=0 0 1200 900, 

%\begin{figure*}
%\begin{center}
%\subfloat[Original]{\includegraphics[width=0.5\linewidth]{"./images/image_22_Original.png"}}
%\subfloat[XLSR]{\includegraphics[width=0.5\linewidth]{"./images/image_22_XLSR.png"}} \\
%\subfloat[ESPCN]{\includegraphics[width=0.5\linewidth]{"./images/image_22_ESPCN.png"}}
%\subfloat[FSRCNN]{\includegraphics[width=0.5\linewidth]{"./images/image_22_FSRCNN.png"}} \\
%\end{center}
%   \caption{Div2K Image No: 0867. Note false colors and staircase gradient effect on quantized ESPCN and FSRCNN output}
%\label{fig:Div2k867}
%\end{figure*}

\section{Conclusion}

In conclusion, we proposed a real-time single image super resolution method driven by the hardware constraints of the Mobile AI 2021 challenge, although it is driven by the target hardware, the resulting model is very efficient in terms of runtime and model parameters. We believe it can run in many mobile hardware with high performance. The proposed model easily surpasses many reported PSNR results of famous FSRCNN and ESPCN models and it even reaches VDSR in most of the public datasets and surpasses its performance on Div2K validation set though it has 30x fewer parameters. Furthermore, the proposed architecture is very robust to uint8 to quantization and only 0.28dB PSNR drop is experienced when compared with float16/32 model. This property of the model makes it a really good candidate for many mobile devices.

%\FloatBarrier
{\small
\bibliographystyle{ieee_fullname}
\bibliography{cvpr}
}

\end{document}